\documentclass[conference]{IEEEtran}
\IEEEoverridecommandlockouts
\usepackage{cite}
\usepackage[hyphens]{url}
\usepackage{multirow}
\usepackage{amsmath,amssymb,amsfonts}
\usepackage{algorithmic}
\usepackage{graphicx}
\usepackage{textcomp}
\usepackage{xcolor}
\usepackage[colorlinks,urlcolor=blue,linkcolor=blue,citecolor=blue]{hyperref}
\def\BibTeX{{\rm B\kern-.05em{\sc i\kern-.025em b}\kern-.08em
    T\kern-.1667em\lower.7ex\hbox{E}\kern-.125emX}}
\begin{document}

\title{A Data-Driven Approach to Enhancing Gravity Models for Trip Demand Prediction\\
\thanks{This material is based upon work supported by the NASA Aeronautics Research Mission Directorate (ARMD) University Leadership Initiative (ULI) under cooperative agreement number 80NSSC23M0059. This research was also partially supported by the U.S. National Science Foundation through Grant No. 2317117 and Grant No. 2309760.}
}

\author{\IEEEauthorblockN{ Kamal Acharya}
\IEEEauthorblockA{\textit{Department of Information Systems} \\
\textit{University of Maryland Baltimore County}\\
Baltimore, MD, US \\
kamala2@umbc.edu}
\and
\IEEEauthorblockN{ Mehul Lad}
\IEEEauthorblockA{\textit{Department of Information Systems} \\
\textit{University of Maryland Baltimore County}\\
Baltimore, MD, US \\
du72811@umbc.edu}
\and
\IEEEauthorblockN{ Liang Sun}
\IEEEauthorblockA{\textit{Department of Mechanical Engineering} \\
\textit{Baylor University}\\
Waco, TX, US \\
liang\_sun@baylor.edu}
\and
\IEEEauthorblockN{ Houbing Song}
\IEEEauthorblockA{\textit{Department of Information Systems} \\
\textit{University of Maryland Baltimore County}\\
Baltimore, MD, US \\
songh@umbc.edu}

}

\maketitle

\begin{abstract}
Accurate prediction of trips between zones is critical for transportation planning, as it supports resource allocation and infrastructure development across various modes of transport. Although the gravity model has been widely used due to its simplicity, it often inadequately represents the complex factors influencing modern travel behavior. This study introduces a data-driven approach to enhance the gravity model by integrating geographical, economic, social, and travel data from the counties in Tennessee and New York state. Using machine learning techniques, we extend the capabilities of the traditional model to handle more complex interactions between variables. Our experiments demonstrate that machine learning-enhanced models significantly outperform the traditional model. Our results show a 51.48\% improvement in $R^2$, indicating a substantial enhancement in the model’s explanatory power. Also, a 63.59\% reduction in Mean Absolute Error (MAE) reflects a significant increase in prediction accuracy. Furthermore, a 44.32\% increase in Common Part of Commuters (CPC) demonstrates improved prediction reliability.  These findings highlight the substantial benefits of integrating diverse datasets and advanced algorithms into transportation models. They provide urban planners and policymakers with more reliable forecasting and decision-making tools.
\end{abstract}


\section{Introduction}
Transportation planning plays a crucial role in ensuring efficient movement of people and goods across various modes of transport. Accurate predictions of trips between geographic zones are foundational to this process, supporting decisions related to infrastructure development, public transit scheduling, and urban growth management~\cite{hasnine2021effects}. One of the most widely used models for predicting trip distributions is the gravity model, which, despite its simplicity, has been applied across different domains for several decades~\cite{anderson2011gravity}. 
With the increasing complexity of trip patterns and the availability of detailed datasets, the need for sophisticated prediction models has become critical~\cite{sanayei2022model}.

While the gravity model~\cite{WILSON1967253} is valuable for its ease of use and interpretability, it has notable limitations, particularly in its ability to account for non-linear relationships and the influence of various socio-economic factors. With the rise of data-driven approaches, machine learning has emerged as a powerful tool to enhance the predictive capabilities of traditional models. Recent advances in data availability, including high-resolution geographic, economic, and travel behavior datasets, offer new opportunities to refine these models and improve their performance.

This study addresses the critical need to enhance trip prediction models by proposing a data-driven extension to the traditional gravity model. We incorporate multiple datasets spanning geographical, economic, social, and travel patterns from counties in two states, Tennessee (TN) and New York (NY) into a machine learning enhanced framework. By doing so, we aim to improve the accuracy and flexibility of trip predictions, offering a more robust tool for transportation planners. 

The key contributions of this paper are:

\begin{itemize} 

    \item First, we integrate a diverse array of datasets to enhance the traditional gravity model’s ability to capture complex, non-linear relationships between variables.
    \item Second, we apply machine learning techniques to improve the accuracy and scalability of trip demand predictions.
    \item Third, we validate the enhanced model with real-world data from counties in TN and NY, achieving up to 50\% improvement in predictive accuracy over traditional methods.

\end{itemize} 

The remainder of this paper is organized as follows: Section~\ref{sec:related_work} presents the Related Works, summarizing prior research. Section~\ref{sec:methodology} details the Methodology, while Section~\ref{sec:evaluation} covers the Evaluation and Discussion. Finally, Section~\ref{sec:conclusion} provides the Conclusion.

\section{Related Works}
\label{sec:related_work}
Transportation planning has long relied on predictive models to forecast trip distributions between geographic zones, with the gravity model being one of the most widely used approaches. The gravity model, first adapted from Newton’s law of gravitation, estimates the flow of trips between two locations based on their population sizes and the distance between them~\cite{erlander1990gravity}. The traditional gravity model is expressed as:
\begin{equation}
    T_{ij} = \frac{P_i \cdot P_j}{d_{ij}^\beta}
\end{equation}
where \(T_{ij}\) is the interaction between the locations \(i\) and \(j\), \(P_i\) and \(P_j\) represent the "masses" of the two locations, \(d_{ij}\) is the distance between them, and \(\beta\) is the distance decay parameter, which reflects how interaction decreases with increasing distance.

Despite its widespread application in urban planning, public transit allocation, and infrastructure development, the gravity model has inherent limitations, particularly when dealing with the complexities of modern transportation systems~\cite{fotheringham1989spatial}. These limitations stem primarily from its inability to capture non-linear relationships and account for various socio-economic factors that influence travel behavior.

Over time, the gravity model has been enhanced with modifications to better reflect real-world complexities. These include adjustments for population and distance effects. For instance, the inclusion of exponents on the population terms (\(P_i^\lambda, P_j^\alpha\)) allows for modifications based on factors such as income or infrastructure that influence interaction beyond population size. Similarly, the distance term (\(d_{ij}^\beta\)) can be tailored to reflect varying distance decay effects, such as travel~\cite{jung2008gravity}. The generalized modified equation is as follows:

\begin{equation}
T_{ij} = k \frac{P_i^\lambda \cdot P_j^\alpha}{d_{ij}^\beta}
\end{equation}
where \(k\) is a scaling constant that adjusts the model for different magnitudes of flow, such as daily or monthly movements.

Another significant expansion to the gravity model involves considering the spatial pattern of origins and destinations, commonly referred to as the spatial structure~\cite{haynes2020gravity}. Spatial structure encapsulates the arrangement and distribution of locations in space and their influence on interaction patterns. In the expanded model, spatial interaction (\(T_{ij}\)) is formulated as a function of three distinct vectors:

\begin{equation}
T_{ij} = f(O_i, D_j, S_{ij})
\end{equation}
where:
\begin{itemize}
    \item \(O_i\): A vector of origin attributes, representing the characteristics of the originating location (e.g., population, economic output, or accessibility);
    \item \(D_j\): A vector of destination attributes, representing the characteristics of the destination location (e.g., attractiveness, size, or infrastructure);
    \item \(S_{ij}\): A vector of separation attributes, capturing the effects of spatial separation, such as distance, travel costs, and intervening opportunities.
\end{itemize}

Researchers have attempted to improve the accuracy of the gravity model by incorporating additional variables, such as travel cost, income disparities, and land-use characteristics~\cite{buliung2006gis}. Although these modifications improve the model’s performance in certain contexts, they often fail to address its rigidity when applied to heterogeneous datasets or complex travel patterns. As transportation systems grow more data-rich, the need for more sophisticated, adaptable models has become evident.

Recent advancements in data availability and machine learning offer new opportunities to overcome the gravity model’s limitations. Machine learning techniques, such as random forests, neural networks (NNs), and support vector machines, are particularly suited to handling non-linear relationships and high-dimensional interactions between variables ~\cite{xie2020urban}. These methods have demonstrated success in various transportation contexts, improving the predictive accuracy of trip demand models. For instance, Ref.~\cite{liyanage2022ai} applied machine learning to bus passenger demand forecasting, significantly improving model performance over traditional methods. Similarly, AI-based models have shown promise in predicting ride-hailing demand and traffic congestion ~\cite{chen2021examining}~\cite{deb2019travel}.

However, while machine learning has been applied to enhance predictive models in transportation, few studies have specifically focused on integrating these techniques with the gravity model. Ref.~\cite{simini2021deep} made strides in this direction by introducing a deep gravity model for mobility flow generation based on geographic data, but comprehensive studies that incorporate diverse, multi-dimensional datasets into gravity models remain limited. Refs.~\cite{zhao2021mdlf} and~\cite{yan2020using} demonstrated the potential of combining machine learning with detailed geographical and socio-economic data for trip destination prediction. Nonetheless, there remains a need for further exploration of how machine learning can specifically address the traditional gravity model’s predictive limitations.

This study contributes to the literature by proposing a machine-learning-enhanced gravity model that integrates a wide range of geographical, economic, social, and travel pattern data. By doing so, it aims to improve the predictive accuracy of trip distributions, providing a more robust tool for transportation planners.

\section{Methodology}
\label{sec:methodology}

\begin{figure}[htbp]
\centerline{\includegraphics[width=\columnwidth]{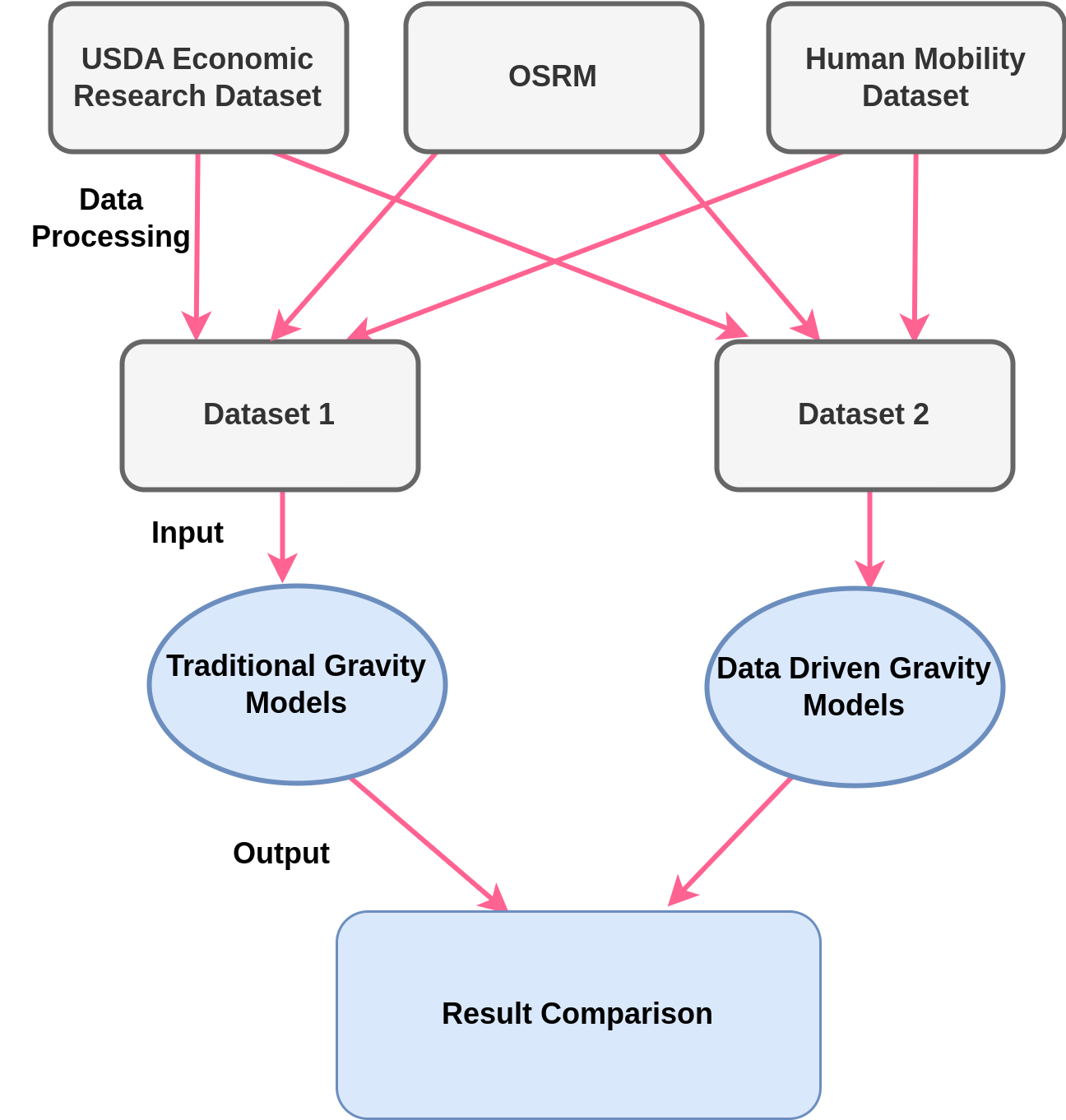}}
\caption{Research Process}
\label{fig:researchMethodology}
\end{figure}

The methodology consists of two phases: first, the implementation of the traditional gravity model as a baseline, and second, the development of a machine-learning-enhanced gravity model that accounts for non-linear relationships between variables as depicted in \autoref{fig:researchMethodology}.

\renewcommand{\arraystretch}{1.25} 
\begin{table*}[h!]
\caption{Description of Feature Categories}
\label{tab:features}
\centering
\begin{tabular}{|p{3cm}|p{13cm}|}
\hline
\textbf{Category (Numbers)} & \textbf{Description} \\ 
\hline
\textbf{Land Use Counts (7)} & F1: Natural, F2: Agricultural, F3: Residential, F4: Commercial, F5: Public, F6: Industrial, and F7: Military purposes. \\ 
\hline
\textbf{Points of Interest (8)} & F8: Education, F9: Commerce, F10: Public Services, F11: Healthcare, F12: Recreation, F13: Heritage, F14:Transport, and F15: Miscellaneous. \\ 
\hline
\textbf{Roads (3)} & F16: Highways, F17: Roadways, and F18: Streets. \\ 
\hline
\textbf{Terminals (1)} & F19: Includes the total count of airports, railway stations, bus stations, etc. \\ 
\hline
\textbf{Structures (1)} & F20: Total building count within each county. \\ 
\hline
\textbf{Economic Features (5)} & F21: Unemployment percentage rate, F22: Median Household Income, F23: \% of State Median Household Income, F24: Percentage of people in poverty, and F25: Percentage of children (0-17) in poverty. \\ 
\hline
\textbf{Education (1)} & F26: Includes percentage of population completing college. \\ 
\hline
\textbf{Population (1)} & F27: Provides population count for each county. \\ 
\hline
\end{tabular}
\end{table*}

\subsection{Study Area}

We selected two states, TN and NY, as the study area. TN has been chosen due to its diverse geographic and demographic composition, making it an ideal region for studying and modeling trip demand patterns. Similarly, NY has been selected as a complementary area of focus due to its high-density urban environments and extensive transportation networks. Together, TN and NY offer distinct yet complementary perspectives—one rooted in regional and rural contexts and the other in densely populated metropolitan areas—enabling a more comprehensive analysis of trip demand across varying geographical and demographic settings. 

At the geographical level, our research focuses on modeling trip demand across counties, which serve as stable administrative subdivisions. Counties provide a consistent framework for data collection and analysis. Although their boundaries generally remain fixed, rare administrative adjustments or annexations can cause changes. However, such adjustments are infrequent, and counties typically retain their boundaries despite population changes, ensuring the continuity of demographic and economic analysis across different time periods. This dual focus on TN and NY allows us to explore mobility solutions in both regional and urban contexts, enriching the scope and applicability of our findings. 


\subsection{Data Collection}
The data for this study was collected from various sources to ensure the inclusion of key factors influencing trip distribution.
The multiscale dynamic human mobility flow dataset~\cite{kang2020multiscale} provides extensive spatiotemporal data on U.S. population movements since January 1, 2019. Compiled from anonymized mobile phone location data sourced by SafeGraph\footnote{SafeGraph: \url{https://www.safegraph.com/}}, this dataset documents origin-to-destination (O-D) population flows at the census tract level on daily bases. Its detailed granularity and temporal coverage make it a valuable resource for applications in transportation planning. For this study, the daily dataset from March 15, 2021, to April 15, 2021, was extracted. Open Source Routing Machine (OSRM)~\cite{luxen-vetter-2011} was utilized to obtain geographical features for counties and also to find the distance and time of trip between those counties. The data related to economy, population and education were sourced from the USDA Economic Research Service\footnote{USDA: \url{https://www.ers.usda.gov/data-products/county-level-data-sets}}. All datasets underwent cleaning, standardization, and were merged using county-level identifiers. Missing values were imputed using median substitution, and numerical variables were scaled to facilitate comparability across features. Ultimately, 27 features were prepared for each county, under eight different categories, as shown in \autoref{tab:features}.


We compile the two sets of datasets by merging the mobility dataset with the others. 
\begin{itemize}
    \item Dataset 1 is based on the original gravity model input, which only considers the population and distance of origin and destination counties as the input, along with the time of travel between the two counties. 
    \item Dataset 2 contains the other geographical, economic and social features along with the previous features.
\end{itemize}

\renewcommand{\arraystretch}{1.25} 

\begin{table*}[ht]
\centering
\caption{Comparison of Gravity Models for TN and NY with Percentage Improvement}
\label{tab:Results}
\begin{tabular}{|c|c|c|c|c|c|c|c|}
\hline
\textbf{Model} & \textbf{Metric} & \multicolumn{3}{c|}{\textbf{TN}} & \multicolumn{3}{c|}{\textbf{NY}} \\
\cline{3-8}
 &  & \textbf{Traditional} & \textbf{Data-Driven} & \textbf{\% Improvement} & \textbf{Traditional} & \textbf{Data-Driven} & \textbf{\% Improvement} \\
\hline
\textbf{Neural Networks} & $MAE$ & 0.1026 & 0.0622 & 39.38\% & 0.0879 & 0.0320 & 63.59\% \\
                         & $R^2$ & 0.8274 & 0.9406 & 13.68\% & 0.6444 & 0.9762 & 51.48\% \\
                         & $CPC$ & 0.7206 & 0.8412 & 16.73\% & 0.6295 & 0.9085 & 44.32\% \\
\hline
\textbf{Random Forest}   & $MAE$ & 0.0420 & 0.0402 & 4.29\% & 0.0255 & 0.0208 & 18.43\% \\
                         & $R^2$ & 0.9746 & 0.9781 & 0.36\% & 0.9755 & 0.9874 & 1.22\% \\
                         & $CPC$ & 0.9121 & 0.9169 & 0.53\% & 0.9291 & 0.9456 & 1.78\% \\
\hline
\textbf{Gradient Boosting} & $MAE$ & 0.1070 & 0.0957 & 10.56\% & 0.0620 & 0.0527 & 15\% \\
                           & $R^2$ & 0.8814 & 0.9180 & 4.15\% & 0.9395 & 0.9654 & 2.75\% \\
                           & $CPC$ & 0.7554 & 0.7894 & 4.50\% & 0.8231 & 0.8631 & 4.86\% \\
\hline
\end{tabular}
\end{table*}

\subsection{Models Implemented}
We have implemented various machine learning models.


\subsubsection{Random Forest}
We implemented a Random Forest model to leverage its capability of handling high-dimensional data and capturing non-linear relationships through ensemble learning. A randomized search cross-validation (RandomizedSearchCV) was employed to tune hyperparameters, optimizing the model’s performance. The hyperparameter space included the number of trees (n\_estimators), maximum tree depth (max\_depth), minimum samples required to split a node (min\_samples\_split), minimum samples required for a leaf node (min\_samples\_leaf), and the number of features considered for splitting (max\_features). The randomized search, performed over 20 iterations with five-fold cross-validation, selected the optimal combination of hyperparameters based on minimizing mean absolute error (MAE). 

\subsubsection{Deep Neural Network}
We implemented a deep neural network model for predicting population flow, and optimized its performance using Optuna. The model architecture featured five fully connected layers with ReLU activation functions, conditional batch normalization (applied based on batch size), and dropout regularization for preventing overfitting. To fine-tune the model, Optuna was used to optimize key hyperparameters, including the learning rate, dropout rate, and batch size. The search space explored learning rates between 1e-5 and 1e-3, dropout rates from 0.1 to 0.5, and batch sizes of 16, 32, and 64. The objective was to minimize the validation MAE. The Adaptive Moment Estimation (ADAM) optimizer, with a final tuned learning rate of 0.00097, was paired with L1 loss to handle the regression task. The optimal hyperparameters were a learning rate of 0.00097, a dropout rate of 0.111, and a batch size of 64.

\subsubsection{Gradient Boosting}
We utilized a Gradient Boosting Regressor (GBR) to model the data, optimizing its performance through randomized hyperparameter tuning. The hyperparameters tuned included the number of estimators (n\_estimators), learning rate, maximum tree depth (max\_depth), minimum samples required to split a node (min\_samples\_split), minimum samples per leaf (min\_samples\_leaf), and subsampling rate (subsample). A randomized search cross-validation (RandomizedSearchCV) was conducted across 10 parameter combinations with two-fold cross-validation to minimize computation time. The best model configuration identified a subsample of 0.9, 500 estimators, a learning rate of 0.05, a maximum depth of 5, with a minimum of 2 samples for splitting and 2 samples per leaf. The optimized model was then trained on the full training dataset, and its performance was evaluated on the test set.





\renewcommand{\arraystretch}{1.25}
\begin{table*}[ht]
\centering
\caption{Top 10 Features for Models in TN and NY}
\label{tab:TopFeatures}
\begin{tabular}{|c|c|l|l|}
\hline
\textbf{State} & \textbf{Model} & \textbf{Ranked Features (1–5)} & \textbf{Ranked Features (6–10)} \\
\hline
\multirow{3}{*}{TN} & Neural Networks & Distance, Time, F27-D, F27-O, F26-D & F4-O, F4-D, F25-O, F17-D, F20-D \\
                    & Random Forest   & Time, F11-D, F14-O, F26-D, F27-D & F16-D, Distance, F18-O, F5-O, F8-O \\
                    & Gradient Boosting & Time, F26-D, F16-D, F27-D, F14-O & F11-D, F17-D, Distance, F8-O, F18-O \\
\hline
\multirow{3}{*}{NY} & Neural Networks & Time, Distance, F12-D, F5-D, F20-O & F5-O, F10-O, F8-D, F10-D, F14-D \\
                    & Random Forest   & Time, Distance, F22-D, F8-D, F21-D & F26-D, F13-D, F23-D, F8-O, F19-D \\
                    & Gradient Boosting & Time, Distance, F26-D, F8-D, F21-D & F20-O, F23-D, F16-O, F16-D, F4-D \\
\hline
\end{tabular}
\end{table*}

\section{Evaluation and Discussion}
\label{sec:evaluation}

In this study, we utilized three evaluation metrics to assess the performance of our model: R-squared ($R^2$),  MAE, and Common Part of Commuters (CPC). The mathematical background for each of these metrics is as follows.

\subsection{R-squared}

$R^2$ also known as the coefficient of determination, measures the proportion of variance in the dependent variable that is explained by the independent variables in the model. A $R^2$ value closer to 1 indicates that the model explains a higher proportion of the variance, implying better fit. Mathematically, it is expressed as:

\begin{equation}
    R^2 = 1 - \frac{\sum_{i=1}^{n} (y_i - \hat{y}_i)^2}{\sum_{i=1}^{n} (y_i - \bar{y})^2}
\end{equation}
where:
\begin{itemize}
    \item \(y_i\) is the actual value,
    \item \(\hat{y}_i\) is the predicted value,
    \item \(\bar{y}\) is the mean of the actual values, and
    \item \(n\) is the number of data points.
\end{itemize}

\subsection{Mean Absolute Error}

MAE is a metric that measures the average magnitude of errors in the model’s predictions, without considering their direction (i.e., positive or negative). It is given by the formula:

\begin{equation}
    MAE = \frac{1}{n} \sum_{i=1}^{n} |y_i - \hat{y}_i|
\end{equation}
where:
\begin{itemize}
    \item \(y_i\) is the actual value,
    \item \(\hat{y}_i\) is the predicted value, and
    \item \(n\) is the number of data points.
\end{itemize}

\subsection{Common Part of Commuters}

CPC, also known as the Sørensen-Dice index, is a metric used to measure the similarity between the generated flows (\(y_g\)) and real flows (\(y_r\)) in flow generation models. CPC ranges from 0 to 1, where 1 indicates a perfect match between the generated and real flows, and 0 signifies no overlap between them. When the total generated outflow equals the real total outflow, the CPC metric becomes equivalent to the accuracy of the model, measuring the fraction of trips assigned to the correct destination. CPC is computed as:

\begin{equation}
    CPC = \frac{2 \sum_{i,j} \min(y_g(l_i, l_j), y_r(l_i, l_j))}{\sum_{i,j} y_g(l_i, l_j) + \sum_{i,j} y_r(l_i, l_j)}
\end{equation}
where:
\begin{itemize}
    \item \(y_g(l_i, l_j)\) represents the generated flow from location \(l_i\) to location \(l_j\),
    \item \(y_r(l_i, l_j)\) represents the real flow from location \(l_i\) to location \(l_j\), and
    \item The summations are over all possible origin-destination pairs \((i, j)\).
\end{itemize}

\autoref{tab:Results} shows a comparison of gravity models for trip demand prediction across TN and NY, highlighting the significant advantages of data-driven approaches. NNs demonstrate the most notable improvement, with the data-driven model reducing the MAE by more than half and substantially increasing \(R^2\), showcasing the effectiveness of non-linear models when incorporating more complex, data-driven features. Random Forest exhibits robust performance in both traditional and data-driven approaches, with the data-driven model slightly improving MAE and \(R^2\), emphasizing the consistency of ensemble methods in capturing complex relationships. Gradient Boosting also achieves notable enhancements, with reductions in MAE and increases in \(R^2\), particularly in the data-driven model, suggesting its suitability for predicting trip demand across varying contexts. Overall, the data-driven models consistently outperform traditional methods, with NNs leading the improvements, especially in high-density urban areas like NY.

\autoref{tab:TopFeatures} presents the top 10 ranked features for the data-driven models in TN and NY for trip demand prediction. SHapley Additive exPlanations (SHAP)  \cite{lundberg2017unified}, a method to interpret machine learning models by quantifying the contribution of each feature, is used to identify the top 10 impactful features for Neural Network outputs while feature importance analysis is used for Random Forest and Gradient Boosting. In TN, for NNs, Random Forest, and Gradient Boosting, key features include distance, time, population count (F27-O), and education (F26-D). These highlight the importance of geographical and educational indicators in regional contexts. For NY, similar features such as distance, time, and population count remain critical across all models, but additional features like public services (F10-D), transportation facilities (F19-D), and commerce-related factors (F9-D) gain prominence, reflecting the urban environment's complexity.

\begin{figure}[htbp]
\centering
\includegraphics[width=0.95\columnwidth]{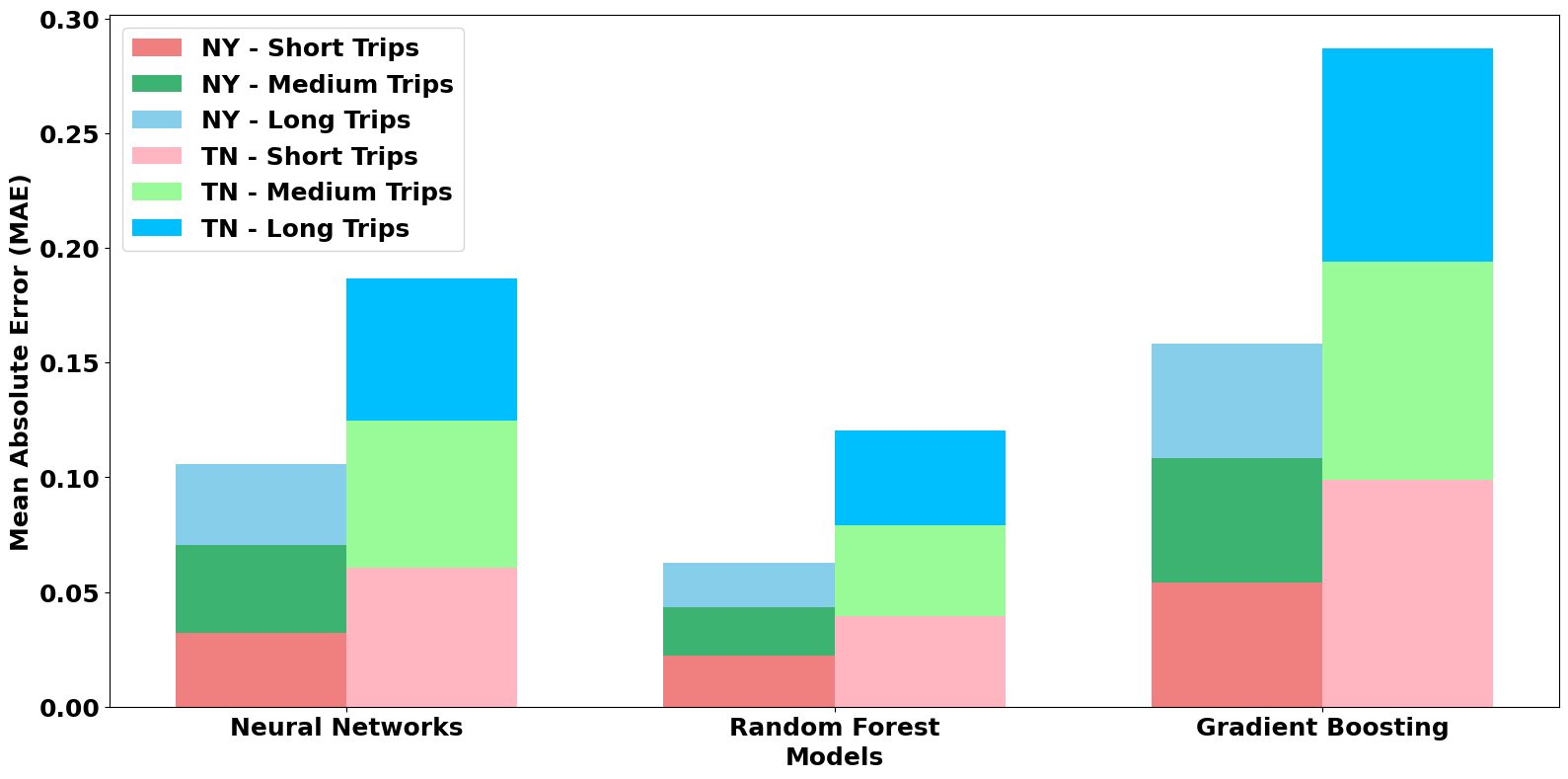}
\caption{MAE Comparison of Models for NY and TN by Trip Distance }
\label{fig:results}
\end{figure}

We then performed trip segmentation analysis by categorizing trips into Short (up to 61.78 miles), Medium (61.79 to 171.44 miles), and Long (above 171.44 miles) based on distance thresholds derived from the dataset's 33rd and 66th percentiles, ensuring balanced trip distribution. The result of the analysis is shown in \autoref{fig:results}. NNs excel in dense urban settings like NY, particularly for Short and Medium Trips, due to their ability to capture non-linear patterns. However, they struggle with Long Trips in TN, where data sparsity poses challenges. Random Forest demonstrates consistent performance across all trip segments and geographies, showcasing its adaptability to diverse data distributions. Gradient Boosting performs well in NY for Medium and Long Trips but falters in TN, reflecting the need for model optimization in regional contexts. 

\begin{figure}[htbp]
\centering
\includegraphics[width=0.95\columnwidth]{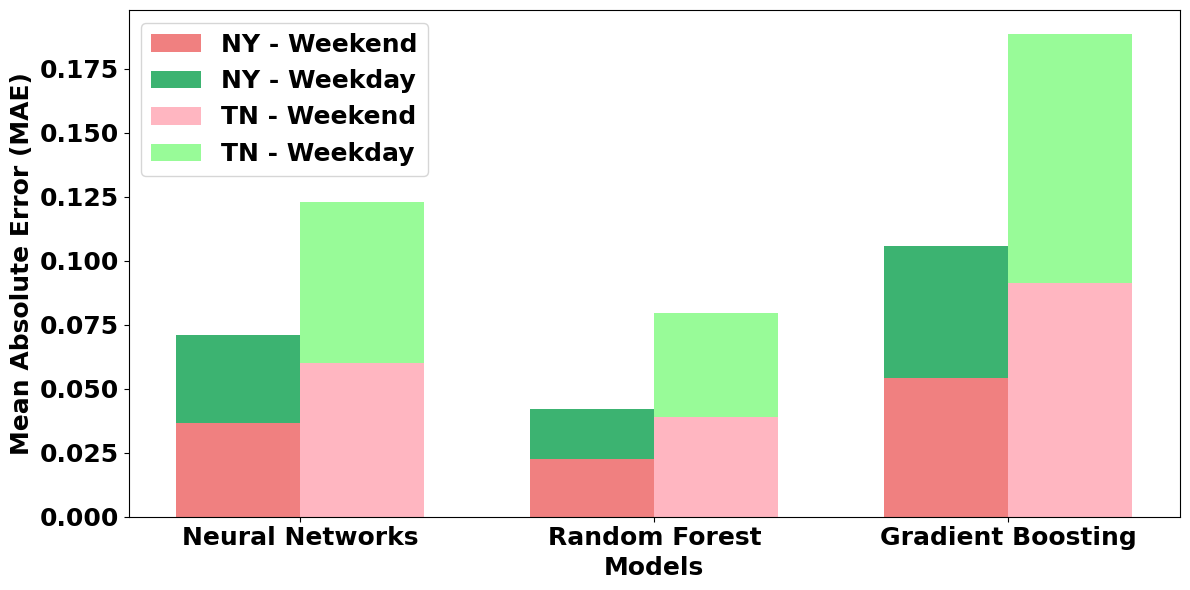}
\caption{MAE Comparison of Models for NY and TN by Days}
\label{fig:days}
\end{figure}

\autoref{fig:days} compares MAE of NNs, Random Forest, and Gradient Boosting models across NY and TN, with distinctions between weekday and weekend travel patterns. NNs show relatively balanced performance between weekday and weekend predictions in both NY and TN, though the errors are noticeably higher in TN for both scenarios, suggesting their limitations in handling sparsity and variability in data from a less dense region. Random Forest consistently demonstrates the lowest MAE among all models, both in NY and TN, across weekdays and weekends. This reflects its robustness in adapting to varying data characteristics, including urban (NY) and rural (TN) environments. However, a slight increase in MAE for TN compared to NY indicates that even Random Forest is not immune to the challenges posed by sparse and heterogeneous datasets. Gradient Boosting, on the other hand, exhibits the highest MAE across all settings for TN, particularly on weekdays, where the error is significantly larger than on weekends. This suggests that the model's sensitivity to sparse and less structured data is amplified during periods of higher variability, such as weekday travel. In contrast, its performance in NY is more competitive, with lower errors across both weekdays and weekends, showcasing its strength in environments with richer and more predictable data patterns.

\section{Conclusion}
\label{sec:conclusion}

This study highlights the substantial benefits of integrating machine learning techniques, such as NNs, Random Forests, and Gradient Boosting, into traditional gravity models for trip demand prediction. By leveraging a diverse set of geographical, economic, social, and travel-related features, the data-driven models consistently outperformed traditional approaches, with reductions in MAE ranging from 10\% to over 60\%, and significant improvements in $R^2$ and CPC, showcasing enhanced predictive accuracy and reliability. NNs demonstrated the most notable improvements, particularly in high-density urban environments like New York, achieving an $R^2$ of 0.976 and a CPC increase of over 44\%. The analysis of top-ranked features, including distance, time, population, and economic factors, further underscores the adaptability of machine learning models to varying regional and urban contexts. Future work could explore the incorporating real-time data, such as GPS and social media activity, to enhance dynamic predictions. Additionally, integrating environmental factors, such as weather and air quality, could provide a more comprehensive understanding of their impact on trip demand patterns.

\section{Acknowledgments}
This material is based upon work supported by the NASA Aeronautics Research Mission Directorate (ARMD) University Leadership Initiative (ULI) under cooperative agreement number 80NSSC23M0059. This research was also partially supported by the U.S. National Science Foundation through Grant No. 2317117 and Grant No. 2309760.

\bibliographystyle{IEEEtran}
\bibliography{ref.bib}
\end{document}